\documentclass[10pt,letterpaper]{article}
\usepackage{cvpr}
\usepackage{times}
\usepackage{epsfig}
\usepackage{graphicx}
\usepackage{amsmath}
\usepackage{amssymb}
\usepackage{subfigure}
\usepackage[rightcaption]{sidecap}
\usepackage{algorithmic}
\usepackage{amsopn}
\usepackage{tabularx, booktabs}
\usepackage{mathtools}
\usepackage[ruled,vlined,linesnumbered,lined,boxed,commentsnumbered]{algorithm2e}
\newtheorem{mydef}{Definition}
\usepackage{tabularx, booktabs}
\usepackage{subfig}
\usepackage{latexsym}
\usepackage{comment}
\newtheorem{Lemma}{Lemma}
\newtheorem{Prop}{Proposition}

\cvprfinalcopy 

\begin{document}

\title{Graph Spectral Embedding using the Geodesic Betweeness Centrality}

\author{Shay Deutsch\\
Department of Mathematics  \\
University of California Los Angeles\\
{\tt\small shaydeu@math.ucla.edu}
\and 
Stefano Soatto\\
Department of Computer Science \\
University of California Los Angeles\\
{\tt\small soatto@cs.ucla.edu }
}

\maketitle

\begin{abstract}
We introduce the Graph Sylvester Embedding (GSE), an unsupervised graph representation of local similarity, connectivity, and global structure. GSE uses the solution of the Sylvester equation to capture both network structure and neighborhood proximity in a single representation. Unlike embeddings based on the eigenvectors of the Laplacian, GSE incorporates two or more basis functions, for instance using the Laplacian and the affinity matrix. Such basis functions are constructed {\em not} from the original graph, but from one whose weights measure the centrality of an edge (the fraction of the number of shortest paths that pass through that edge) in the original graph. This allows more flexibility and control to represent complex network structure and shows significant improvements over the state of the art when used for data analysis tasks such as predicting failed edges in material science and network alignment in the human-SARS CoV-2 protein-protein interactome.  
\end{abstract}

\section{Introduction}

Analysis of networks arising in social science, material science, biology and commerce, relies on both local and global properties of the graph that represents the network. Local properties include similarity of attributes among adjacent nodes, and global properties include the structural role of nodes or edges within the graph, as well as similarity among distant nodes. Graph representations based on the eigenvectors of the standard Graph Laplacian have limited expressivity when it comes to capturing local and structural similarity, which we wish to overcome.  

We propose an {\em unsupervised} approach to learn embeddings that exploit both the Laplacian and the affinity matrix combined into a Sylvester equation. However, to better capture local as well as structural similarity, we consider the Laplacian and the affinity matrix  {\em not}  of the original graph, but rather of a modified {\em betweenness centrality graph} derived from it. Centrality is a statistic (a function of) the topology of the graph that is not captured in the original edge weights.  The node embeddings are the Spectral Kernel Descriptors of the solution of the Sylvester equation.

The key to our approach is three-fold: (i) The betweeness centrality of edges, defined as the fraction of shortest paths going through that edge, to capture semi-global structure in the original graph; (ii) The use of both the Laplacian and the affinity matrix of the centrality graph, to modulate the effect of local and global structure; (iii) The use of the Sylvester equation as a natural way of combining these two graph properties into a single basis. Any number of descriptors could be built on this basis, we just choose  the Spectral Kernel Descriptor for simplicity.

Our approach is related to uncertainty principles of graph harmonic analysis, that explore the extent in which signals can be represented simultaneously in the vertex and spectral domains. We extend this approach to characterize the relation between the spectral and non-local graph domain spread of signals defined on the nodes. 

We illustrate the use of  our method in multiple unsupervised learning tasks including node classification in social networks, network alignment in protein-to-protein interaction (PPI) for the human-SARS-CoV-2 protein-protein interactome, and forecasting failure of edges in material science. 

\begin{figure*}
\centering
\includegraphics[width=0.6\columnwidth]{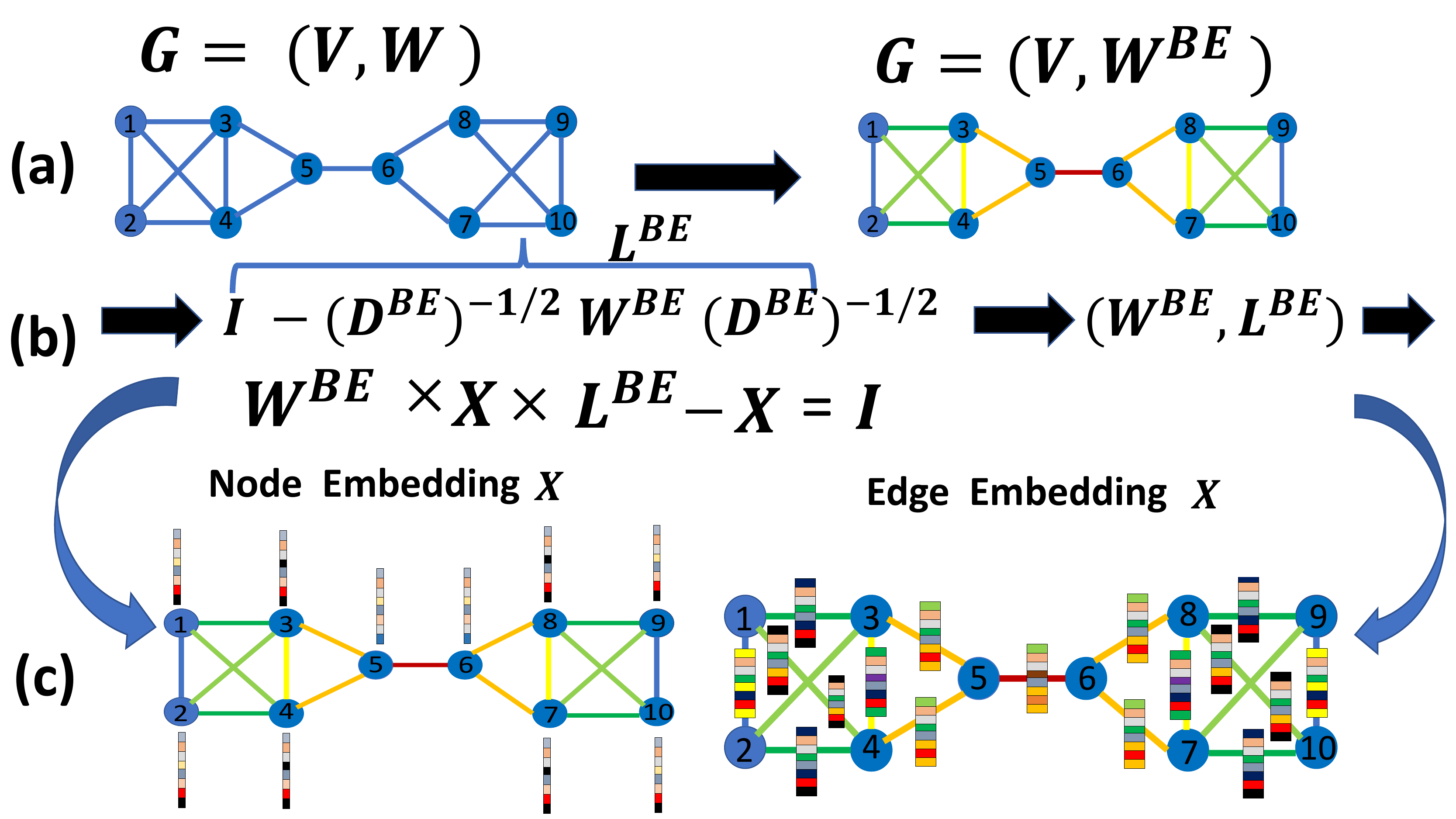}
\caption{Construction of the Graph Sylvester Embedding (GSE): (a) the original graph is transformed into the centrality graph, by replacing $w_{i,j}$ with $w_{i,j}^{BE} $; (b) $\mathbf{W}^{BE}$ and $\mathbf{L}^{BE}$ are combined into a Sylvester equation, whose solution is used to define (c) node embeddings using the Spectral Kernel, from which edge embedding can be constructed. From the color signatures (best seen zoomed in), we can see that nodes that have different structural properties have different descriptors, whereas nodes with similar structural properties have similar descriptors even when they are not adjacent.}  
\label{fig:barbell_graph_illustrate}
\end{figure*}

\section{Related Work}
\label{sec:RelatedWork}


\def\BCG{\bar{\cal G}}

For a graph ${\cal G}= (\cal V, W)$ with vertices in the set $\cal V$ $ = \left \{ 1, 2, . . . , N \right \} $ and edge weights in $\mathbf{\cal{W}} = \left \{ (w_{ij}) (i, j) \in \cal{V}  \right \} $, where $w_{ij}$ denotes the weight of an edge between nodes $i$ and $j$, we measure the {\em centrality} of an edge by the fraction of the number of shortest paths that pass through that edge, called {\em edge betweenness centrality} (EBC). From the betweenness centrality of the edges of ${\cal G}$, we construct a modified graph with the same connectivity of ${\cal G}$, but edges weighted by their EBC value. The result is called edge betweeness centrality graph (BCG), or {\em centrality graph} for short, and denoted by $\BCG$, which measures the importance of edges given the adjacency matrix. 

Graph centrality based measurements, such as vertex centrality, eigenvector centrality \cite{Freeman1977Set}, and edge betweeness centrality \cite{Girvan2002Community} have been widely used in network analysis and its diverse applications \cite{Berthier_2019, CUCURINGU_2016}. 
There is a large body of work that focus on graph embedding methods that, similar to our approach, aims to learn a function $\Phi:\cal V \rightarrow$  $ \mathbb{R}^{m} $ from the graph nodes into a vector space $ \mathbb{R}^{m}$ \cite{UnifyingDeepWalk, grover2016node2vec, Deepwalk, CommunityPreservingNetwork}.  Such methods construct graph representations that are based on either local or structural similarities.

Graph convolutional networks \cite{Kipf:2016, NIPS2016CNNG, velickovic2018graph} are among the most popular methods for graph based feature learning, combining the expressiveness of neural networks with the graph structures. Typically, these methods assume that node features or attributes  are available, while we focus on problems where only the graph network is available. There is also a large body of work that focus on graph embedding methods that, similar to our approach, only assumes that the affinity matrix is given as an input.  Such methods construct graph representations that are based on either local or structural similarities \cite{UnifyingDeepWalk, grover2016node2vec, Deepwalk, CommunityPreservingNetwork}. 

Uncertainty principles in graphs have been explored by \cite{Agaskar_2013,Teke_U_Principles}, extending traditional uncertainty principles of signal processing to more general domains such as irregular and unweighted graphs \cite{GSP_irregular}. \cite{Agaskar_2013} provides definitions of graph and spectral spreads, where the graph spread is defined for an \textit{arbitrary} vertex, and studies to what extent a signal can be localized both in the graph and spectral domains.  

Other related work includes different approaches in manifold learning \cite{LLE,Belkin:2003,Tenenbaum00}, manifold regularization \cite{MDhein,MFD,ZSLwavelets,DeutschM17,ZSL_Isoperimteric,DeutschOM18,ijcai15,SSVM15}, graph diffusion kernels \cite{CoifmanDM} and kernel methods widely used in computer graphics \cite{AubrySC11, HKS} for shape detection. Most methods assume that signals defined over the graph nodes exhibit some level of smoothness with respect to the connectivity of the nodes in the graph and therefore are biased to capture local similarity.

The Sylvester equation was previously employed in a variety of graph mining applications such as network alignment and subgraph matching \cite{Final, Fasten}. Most previous works used the \textit{continuous Sylvester equation} with two input affinity matrices corresponding to the input affinity matrices. We propose a different approach, where we solve the \textit{discrete-time Sylvester equation} using the affinity matrix and its associated Laplacian. 


\section{Preliminaries and Definitions}
\label{sec:Preliminaries}

Consider an undirected, weighted graph $\cal{G}=(V,W)$.
The degree $d(i)$ of a node is the sum of the weights of edges connected to it. 
 The combinatorial graph Laplacian is denoted by $\mathbf{L}$, and defined as $\mathbf{L}=\mathbf{D}-\mathbf{W}$, with  $\mathbf{D}$ the diagonal degree matrix with entries $d_{ii}=d(i)$. The eigenvalues and eigenvectors of $\mathbf{L}$ are denoted by $\lambda_1,\ldots,\lambda_N $ and $\mathbf{\phi}_{1},\ldots,\phi_N$, respectively. The normalized Laplacian is defined as ${\mathbf{L}}_{N} = \mathbf{D}^{-\frac{1}{2} } \mathbf{L} \mathbf{D}^{-\frac{1}{2} }  = \mathbf{I} - \mathbf{D}^{-\frac{1}{2} } \mathbf{W} \mathbf{D}^{-\frac{1}{2} }$ and its real eigenvalues are in the interval $[0,2]$. 
 In this work we use the normalized Laplacian $\mathbf{L}_{N}$, which from now on we refer to as  $\mathbf{L}$ for simplicity. 

\subsection{Centrality Graph}

The edge betweeness centrality (EBC) is defined as 
\begin{equation}
w_{i,j}^{BE} =
\sum_{s \neq t} \frac{ \sigma_{st} (w_{ij})} {\sigma_{st}}
\end{equation} where $\sigma_{st}$ is the number of the shortest distance paths from node $s$ to node $t$ and $\sigma_{st} (w_{ij}) $ is the number of those paths that includes the edge $w_{ij}$. Since we do not use other forms of centrality in this work, we refer to EBC simply as centrality. Accordingly, the (edge betweenness) {\em centrality graph} BCG is defined as $\BCG = (\cal V, W^{BE}) $ which shares the same connectivity but modifies the edges of $\cal G$ to be $w_{i,j}^{BE}$. Formally, the similarity matrix representing the connectivity  of $\BCG$ is 
\begin{equation}
\label{Aff_Bet_graph}
{ ( {\mathbf{W})_{i,j}}^{BE}} = \left\{\begin{matrix}
\sum_{s \neq t} \frac{ \sigma_{st} (w_{ij})} {\sigma_{st}} & \mbox{if}\,\, i\sim~j \, \, \mbox{in} \,\,\,\cal G \\ 
0 & \mbox{else.}
\end{matrix}\right.
\end{equation}
Note that while $\cal G$ and $\BCG$ share the same connectivity, their spectral representation given by the eigensystem of the graph Laplacian $\mathbf{L}^{BE}$, the eigenvalues and their associated eigenvectors  $\lambda^{\mathbf{L}^{BE}}_{1},\ldots,\lambda^{\mathbf{L}^{BE}}_{N} $ and $\mathbf{\phi}^{\mathbf{L}^{BE}}_{1},\ldots,\phi^{\mathbf{L}^{BE}}_{N}$, are rather different.  
Importantly, the eigenvectors $\mathbf{\phi}^{BE}_{1},\ldots,\phi^{BE}_{N}$ provide a different realization of the graph structure in comparison to the eigenvectors of an unweighted graph, that is captured by a different diffusion process around each vertex.  We denote with $\Phi^{\mathbf{W}^{BE}},  \Phi^{\mathbf{L}^{BE}} $ the matrices corresponding to the eigenvector decomposition of  $\mathbf{W}^{BE}, \mathbf{L}^{BE}$ respectively. 

\subsection{Structural and local similarity trade-off Via 
 Total Vertex and Spectral Spreads}
\label{sec:vertex_spectral_spread}

\begin{figure}
\centering
\includegraphics[width=.3\columnwidth]{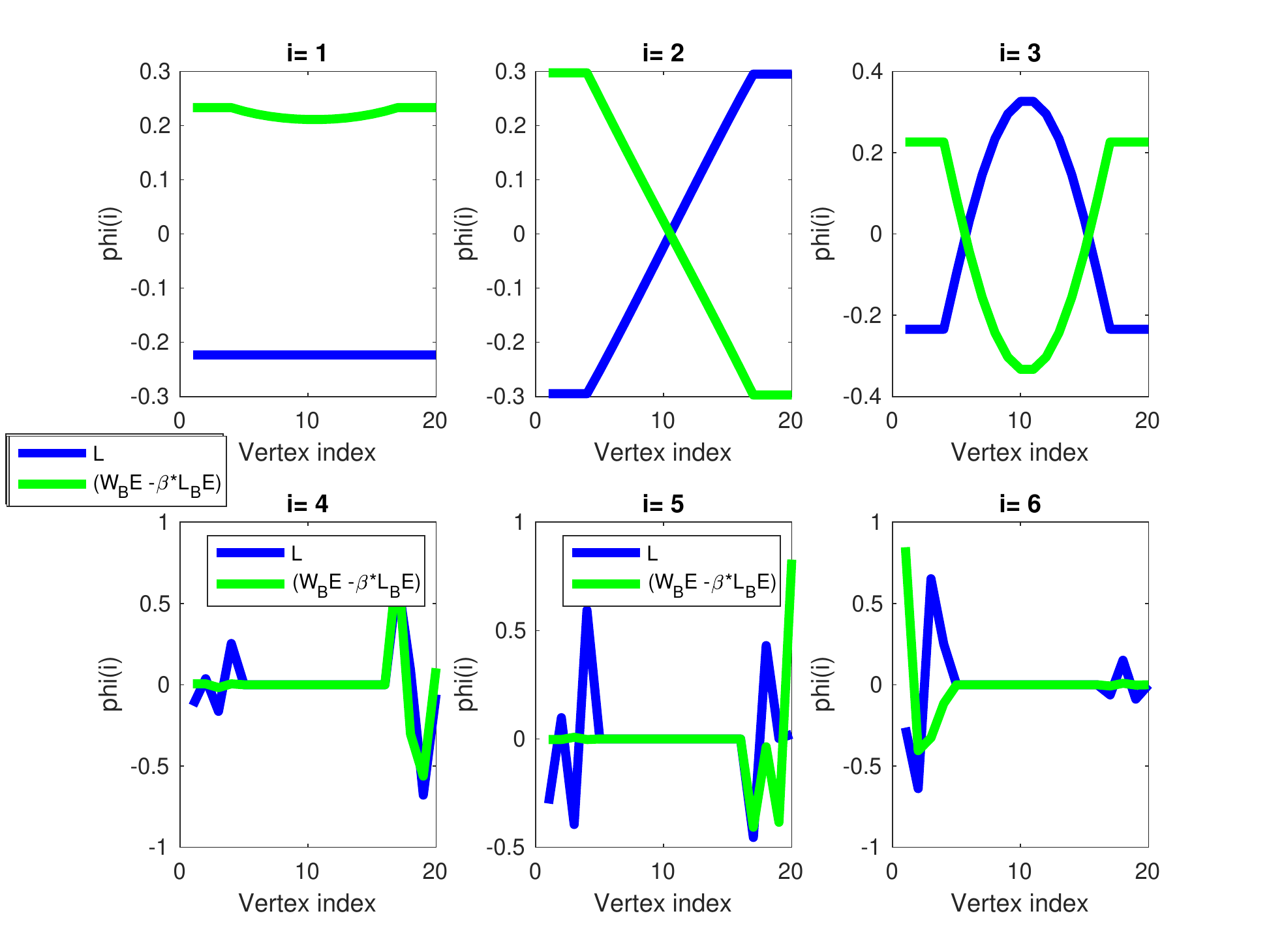}
\includegraphics[width=.3\columnwidth]{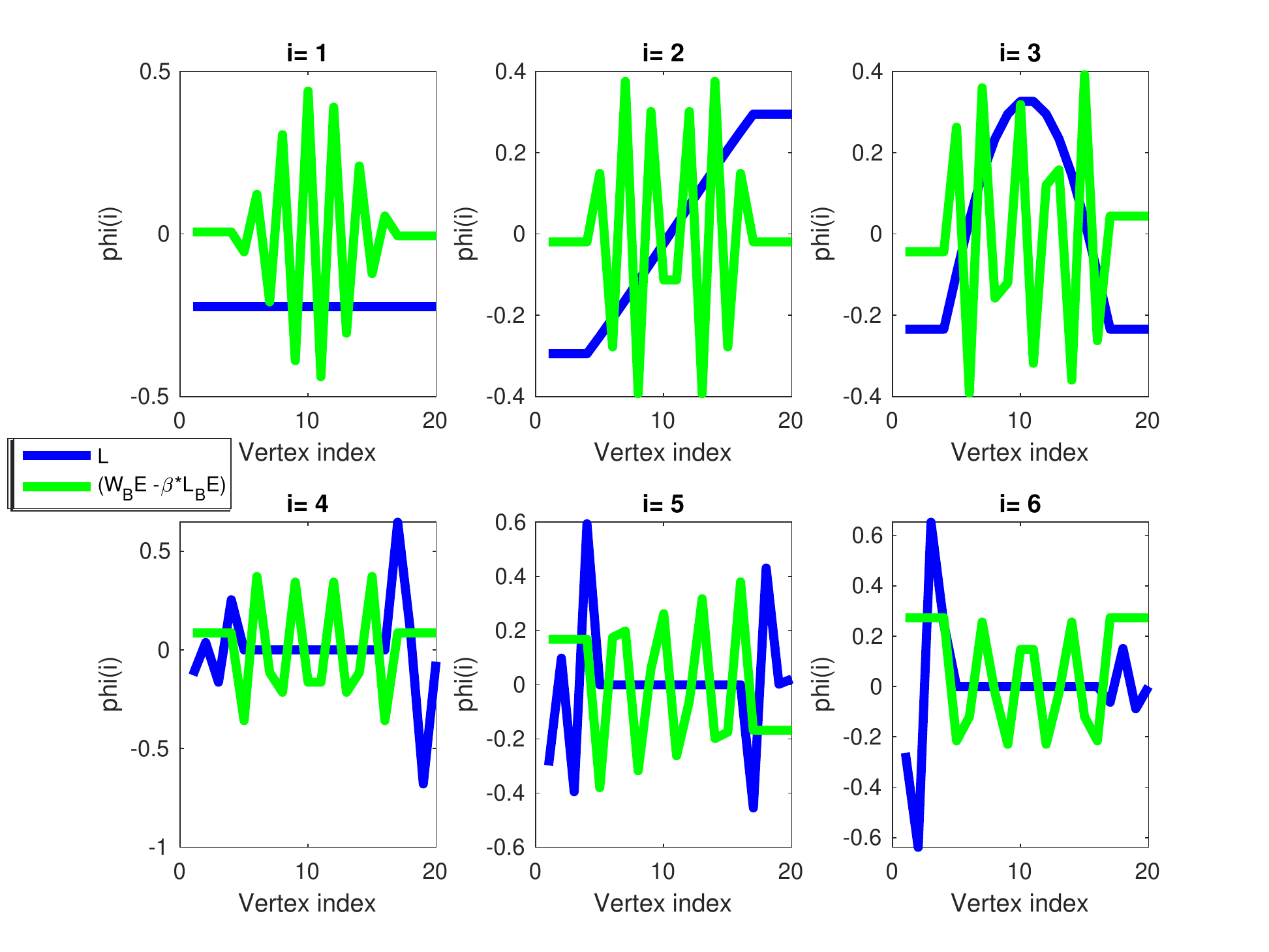}
\caption{The trade-off between local and structural node similarity in the barbell graph, as captured by $\tilde{\mathbf{L}}^{BE}(\beta$) in Figures (a) and (b), showing the eigenvectors associated with the smallest six eigenvalues corresponding to $\tilde{\mathbf{L}}^{BE}(\beta)$ (green color) using  $\beta$ = -200 in (a) and $\beta$ = -0.2 in (b). Eigenvectors of $\mathbf{{L}}^{BE}$are shown in blue color.}
\label{Eigenvectors_Barbell}
\end{figure}

The eigensystem of the Laplacian $\mathbf{L}^{BE}$ and the affinity matrix $\mathbf{W}^{BE}$ provides two alternative basis functions that capture different properties of the graph structure (see Fig. \ref{Eigenvectors_Barbell} for an illustration). 
One way to study the basis functions' different characterizing network structure and local structure is using the vertex and spectral spreads (see their definitions below).  

\begin{mydef}{(Total vertex domain spread)}\\
The global spread of a non zero signal $\mathbf{x}  \in l^{2}(\cal G)$ with respect to a matrix $\mathbf{W}$ (corresponding to an arbitrary affinity matrix) is defined as 
\begin{equation}
 \mathbf{g}_{\mathbf{W}}(\mathbf{x}) =  \frac{1}{ || \mathbf{x} ||^{2}}   \sum_{ i \sim j}   w_{ij}  x(i) x(j) =   \frac{1}{ || \mathbf{x} ||^{2}} \mathbf{x}^{T}\mathbf{W}\mathbf{x}
\end{equation}where $i \sim j$ corresponds to vertices connected in the similarity graph $\mathbf{W}$.  

\end{mydef}

\begin{mydef}{(Spectral spread)  \cite{Agaskar_2013}}\\
The spectral spread of a non zero signal $\mathbf{x}  \in l^{2}(\cal G)$ with respect to a similarity affinity matrix $\mathbf{W}$ is defined as 
\begin{equation}
\label{spectral_spread}
\mathbf{g}_{\mathbf{L}}(\mathbf{x}) =  \frac{1}{ || \mathbf{x} ||^{2}}   \sum_{ l=1 }^{N}  \lambda_{l} | {\hat{x}}(l)|^{2}=    
 \frac{1}{ || \mathbf{x} ||^{2}} \mathbf{x}^{T}\mathbf{L}\mathbf{x}
\end{equation}
where 
\begin{equation}
 {\hat{x}}(l) =  \sum_{i} x(i) \phi^{*}_{l}(i)
\end{equation}is the graph Fourier transform of the signal  $\mathbf{x}  \in l^{2}(\cal G)$ with respect to the eigenvalue $\lambda_{l}$.  
\end{mydef}

The trade-off can be realized by characterizing the domain enclosing all possible pairs of vertex and spectral spreads of the BCG.
\begin{mydef} {Feasibility domain of $\BCG$ }\\
\begin{equation}
\label{feasiblity_doamin}
\mathbb{D}_{ ( s_{    \mathbf{L} },  s_{\mathbf{  \mathbf{W}  } }   )  }=\left \{  (s_{   \mathbf{L}}, s_{\mathbf{  \mathbf{W} } })  | \, g_{\mathbf{L} }(\mathbf{x})  = s_{\mathbf{L} }  ,  g_{  \mathbf{W} }(\mathbf{x})  =   s_{ \mathbf{W}   } 
  \right \}
\end{equation} where $\mathbf{x} \in l^{2}(\cal G)$. 
\end{mydef}

Specifically, searching for the lower boundary of the feasibility domain of the vertex and spectral spreads of $\BCG$ is shown to yield a generalized eigenvalue problem, whose corresponding eigenvectors produce a representation which trades off  local and structural node similarity.  We can write the generalized eigenvalue problem (\ref{generalized eigenvalue problem}) using the matrix pencil $\mathbf{\tilde{L}}(\beta)$ 
\begin{equation}
\mathbf{\tilde{L}}(\beta)\mathbf{x} =  \gamma \mathbf{x}  
\end{equation}where

\begin{equation}
\label{generalized eigenvalue problem}
\mathbf{\tilde{L}}(\beta) = ( \mathbf{W}^{BE}   - \beta \mathbf{L}^{BE}) \mathbf{x} 
\end{equation}
where the eigenvector $\mathbf{x}$ solving (\ref{generalized eigenvalue problem}) is also a minimizer for (\ref{Feasiblility_Domain}) (see the appendix for additional details). 

The scalar $\beta \in \mathbb{R}$  controls the trade-off between the total vertex and spectral spreads, as illustrated in the barbell graph $\tilde{\mathbf{L}}^{BE}(\beta$) in Fig.\ref{Eigenvectors_Barbell} (a) and (b) using $\beta$ = -200, and $\beta$ = -0.2, respectively. Functions colored with green correspond to the eigenvectors of $\tilde{\mathbf{L}}^{BE}(\beta$),  functions colored with blue correspond to the eigenvectors of ${\mathbf{L}}^{BE}$. As can be seen, when  $|\beta|$ is large,  the eigenvectors of $\tilde{\mathbf{L}}^{BE}(\beta$)  reveal structure which is similar to the eigenvectors of ${\mathbf{L}}^{BE}$, while small values of $|\beta|$ produce structure which is similar to those corresponding to ${\mathbf{W}}^{BE}$.\\

\section{Proposed Graph Sylvester Embedding}
\label{sec:Embedding_Sylvester}

We suggest an embedding that goes beyond the scalar modulation of the generalized Laplacian Eq.(\ref{generalized eigenvalue problem}) by using a linear mapping between the subspaces $\mathbf{W}^{BE}$ and  $\mathbf{{L}}^{BE}$. We exploit the fact that the correspondence between the nodes represented by the two basis functions of $\mathbf{W}^{BE}$and $\mathbf{{L}}^{BE}$ is known,  i.e., it is the identity map,  $\pi : {\mathbf{W}}^{BE}\rightarrow {\mathbf{L}}^{BE}$  with $\pi( {\mathbf{W}}^{BE}(:,i)) ={\mathbf{L}}^{BE}(:,i)$ for each $i \in \cal V$ (i.e., the column index corresponding to each node $i \in \cal V$ in ${\mathbf{W}}^{BE}$ is the same in ${\mathbf{L}}^{BE}$).  We propose using the solution of a Sylvester equation as the node feature representation, which is also the mapping between the nodes' representations. The resulting embedding is given by the solution to the discrete Sylvester equation (or the singular value decomposition associated with the solution) will be composed of two hybrid representations associated with the graph networks and local connectivity. The proposed method is coined Graph Sylvester Embedding (GSE). 

The \textit{discrete-time} Sylvester operator $\mathbf{S}(\mathbf{X}) = \mathbf{A} \mathbf{X}\mathbf{B}  - \mathbf{X} $  is used to express the eigenvalues and eigenvectors of $\mathbf{A}$ and $\mathbf{B}$ using a single operator $\mathbf{S}$, where we solve
\begin{equation}
\label{Sylvester}
  \mathbf{S}(\mathbf{X}) = \mathbf{A} \mathbf{X}\mathbf{B}  - \mathbf{X}  = \mathbf{C} 
\end{equation}
using $\mathbf{C} =   \mathbf{I} $,  ($\mathbf{I} \in \mathbb{R}^{N\times N}$ is the identity matrix), $\mathbf{A} = \mathbf{W}^{BE}, \mathbf{B}  = \mathbf{L}^{BE}$.

Let $ \left \{   \phi_{i}^{\mathbf{W}^{BE}} \right \}_{l=1}^{N} $, $ \left \{  \lambda_{l}^{\mathbf{W}^{BE}}  \right \} _{l=1}^{N}$, and $ \left\{  \phi_{i}^{\mathbf{L}^{BE}} \right \}_{l=1}^{N} $, $ \left \{  \lambda_{l}^{\mathbf{L}^{BE}}  \right \} _{l=1}^{N}$ be the corresponding eigenvectors and eigenvalues of $\mathbf{A}$ and $  \mathbf{B}$, respectively.  

We will assume that the eigenvalues of $\mathbf{A}$ and $\mathbf{B}$ satisfy that $\lambda_{i}^{\mathbf{L}^{BE}} \lambda_{j}^{\mathbf{W}^{BE}}\neq -1 $, $\forall i ,j $, hence the operator $\mathbf{S}(\mathbf{X}):\mathbb{R}^{N \times  N   } \rightarrow   \mathbb{R}^{N  \times  N} $ is non-singular and has $N^2$ matrix eigenvalues and eigenvectors. Note that  the Sylvester equation has a unique solution for any $\mathbf{C}$  if and only if $\mathbf{S}$ is non-singular, which occurs if and only if $\lambda_{i}^{\mathbf{L}^{BE}} \lambda_{j}^{\mathbf{W}^{BE}}\neq -1$ $\forall i ,j $. By choosing $ \mathbf{C} = \mathbf{I}$ and ordering the columns in $\mathbf{A}$ and $\mathbf{B}$ based on the identity map, $\pi : \mathbf{A}\rightarrow \mathbf{B}$ we obtain that each column in $\mathbf{X}$ provides a node representation to its corresponding node  $i \in \cal V$. 

\begin{figure*}
\centering
\includegraphics[width=0.6\columnwidth]{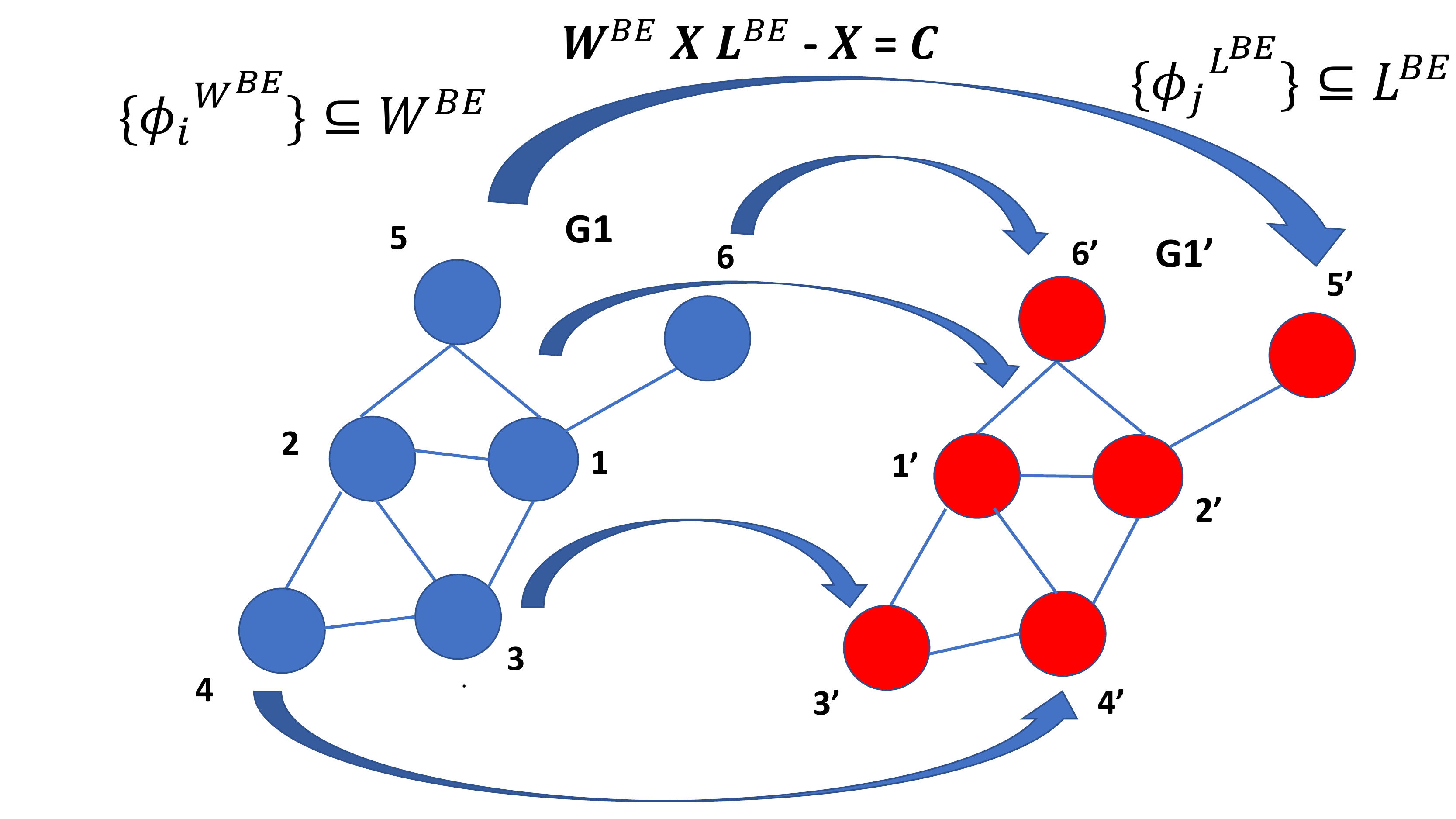}
\caption{In the proposed Graph Sylvester Embedding (GSE), the mapping between two basis functions is used to construct the proposed embedding, exploiting the known correspondence between the two alternative graph representations.}  
\label{fig:Mapping_Graph_illus}
\end{figure*}

Note that we can also express the solution $\mathbf{X}$ to the Sylvester equation Eq.(\ref{Sylvester}) using the Kronecker product $\otimes$:
\begin{equation}
(\mathbf{I}  - \mathbf{B}^{T} \otimes  \mathbf{A})x = c
\end{equation}
where we vectorize $\mathbf{X}$ and $\mathbf{C}$ to obtain its equivalent vector representation  $x$ = vec $(\mathbf{X})$ and $c$ = vec$(\mathbf{C})$. 
Since the Sylvester equation can be written using the Kronecker product, then all known properties of the Kronecker product are carried on to the proposed graph Sylvester embedding. One important property is invariance to permutations. 
\begin{Prop}
\label{permutation_inv}
Suppose that $\mathbf{P} \in \mathbb{R}^{N \times N}$ is a permutation matrix, $\mathbf{\tilde{A}} = \mathbf{P}^{T}\mathbf{A} \mathbf{P}$, and $\mathbf{\tilde{B}} = \mathbf{P}^{T}\mathbf{B} \mathbf{P}$, and assume that $\mathbf{X}, \mathbf{\tilde{X} } $ are two solutions to the discrete-time Sylvester equation Eq.(\ref{eqn:Sylvester}) with the associate matrices  $\mathbf{{A}}, \mathbf{{B}}$ and  $\mathbf{\tilde{A}} ,   \mathbf{\tilde{B}}$, respectively, and $\lambda_{i}^{\mathbf{L}^{BE}} \lambda_{j}^{\mathbf{W}^{BE}}\neq -1 $, $\forall i ,j $, then 
\begin{equation}
\mathbf{\tilde{X}} = \mathbf{P}^{T}\mathbf{X} \mathbf{P}
\end{equation}

\end{Prop}
{Proof:} See Appendix.


The resulting solution of $\mathbf{X}$ can be described using the following conditions. Assume that $ \lambda_{i}^{\mathbf{W}^{BE}}\lambda_{j}^{\mathbf{L}^{BE}}  \neq 1 \, \forall i, j $,  and that the graph representation encoded in $\mathbf{A}$ and $\mathbf{B}$ has the same order based on the identity map, and $ \mathbf{C}=  \mathbf{I}$. Then the solution $\mathbf{X}$ to the discrete-time Sylvester system Eq.(\ref{Sylvester}) is unique where 
\begin{equation}
\label{Sylvester_analytical}
\mathbf{X} = \mathbf{U}^{\mathbf{W}^{BE}}\mathbf{\tilde{C}} (\mathbf{V}^{\mathbf{L}^{BE}})^{T}
\end{equation}
where $\mathbf{\tilde{C}}_{i,j} = 
\frac{   (\phi_{i}^{\mathbf{W}^{BE}})^t \phi_{j}^{\mathbf{L}^{BE}}    }   {\lambda_{i}^{\mathbf{W}^{BE}}\lambda_{j}^{\mathbf{L}^{BE}} - 1}$. 
 See proof in the appendix. 

\textbf{Remark 1}: Eq. (\ref{Sylvester_analytical}) shows the analytical solution to the Sylvester equation. In practice, one can use fast iterative methods  \cite{Large-Scale_Stein}. Also note that under the assumptions that  $ \lambda_{i}^{\mathbf{W}^{BE}}\lambda_{j}^{\mathbf{L}^{BE}}  \neq 1 \, \forall i,j$, the solution $\mathbf{X}$ is unique \cite{Large-Scale_Stein}. We note that Sylvester's equation can be generalized to include multiple terms, thus allowing one to incorporate multiple basis functions. We illustrate the concept with two bases, to capture local and global statistics of the topology of the network, and leave extension to additional bases, that can be specific to particular domains or tasks, for future work.


\textbf{Remark 2}: One can view the solution $\mathbf{X}$ to the Sylvester equation as a polynomial of the matrices $\mathbf{A}$ and $\mathbf{B}$ (\cite{Poly_Sol_Sylvester}), where using higher-order polynomials in $\mathbf{W}^{BE}$, $\mathbf{L}^{BE}$ can be interpreted as imposing smoothness in the embedding space.

\subsection{Spectral Representation of GSE}
By our specific construction of matrices in the Sylvester equation Eq.( \ref{Sylvester_analytical}) we have that the $i$th row of the solution $\mathbf{X}$ captures local and global statistics of the topology of the network with respect to the node $i$. This is an important property which was achieved by our choice of $\mathbf{C} = \mathbf{I}$, and using the same order of the graph representation encoded in $\mathbf{W}^{BE}$ and $\mathbf{L}^{BE}$.  It is now possible to construct graph embedding from $\mathbf{X}$. We employ graph embedding using the spectral decomposition of  $\mathbf{X}$. 

Given the Singular Value Decomposition  $\mathbf{X} = U\Lambda V^{*}$ and equally spaced scales $\left \{ t_{s} \right \}$, we compute a node embedding for each node $i \in \cal G$ using the Spectral Kernel descriptor of $\mathbf{X}$ using 
 \begin{align}
 \label{kernel_signature}
 \psi(\mathbf{x}_{i}, t_{s})  
 =\sum_{l=1}^{m} c_{t_{s}}\mbox{exp}    \left (  -\frac{ ( \mbox{log}(t_{s}) -\mbox{log} ( \lambda_{l}  )  )^{2} } {2\sigma^{2}  }   \right ) ( {\mathbf{u}_{l}  })^{2}  (i)
 \end{align}
where $U = \left \{  \mathbf{u}_{l}\right \} ,  V=\left \{  \mathbf{v}_{l} \right \} $, and $\Lambda  = \left \{   \lambda_{l} \right \}  $ correspond to the left and right singular eigenvectors, and singular values, respectively, $m$ is to the number of the largest eigenvalues and eigenvectors used in the SVD decomposition, and $c_{t_{s}}$are normalizing constants.  \\
 \textbf{Remark:} Note that the spectral kernel descriptor in Eq.(\ref{kernel_signature}) is similar to the WKS descriptor \cite{AubrySC11} proposed to describe the spectral signature of the Laplace- Beltrami operator (LBO). For applications using Sylvester embedding shown in this work, we find the descriptor in Eq.(\ref{kernel_signature}) effective as it weighs the eigenvalues equally and separates the influence between different eigenvalues.

 
 \begin{algorithm}[tb]
   \caption{Graph Sylvester Embedding (GSE)}
   \label{alg:GSE}
\begin{algorithmic}
   \STATE {\bfseries Input:} Graph $\cal{G}=(V,W)$, embedding dimension $m$. equally spaced scales $t_{s}$.   
    \STATE  {\bfseries Output:} Node Embedding $\mathbf{y}_{i}\in \mathbf{R}^{m} $ for each $i \in \cal{G} $ 
   \STATE \textbf{Step 1:} Compute the $\BCG$  using Eq.(\ref{Aff_Bet_graph}). 
   \STATE \textbf{Step 2:} Compute \\ $  {{\mathbf{L}}^{BE}} = 
   \mathbf{I} - ({\mathbf{D}^{BE}})^{-\frac{1}{2} } \mathbf{W}^{BE} ({\mathbf{D}^{BE}})^{-\frac{1}{2} }  $ using $\BCG$  computed in Step 1. 
   \STATE \textbf{Step 3:} Solve the discrete-time Sylvester equation:
   \begin{equation}
   \mathbf{A} \mathbf{X}\mathbf{B}  - \mathbf{X} =  \mathbf{C}
   \end{equation}
    using $ \mathbf{A}  = \mathbf{W}^{BE} $, $ \mathbf{B}=\mathbf{L}^{BE}$, and $\mathbf{C} = \mathbf{I}$  . 
   \STATE \textbf{Step 4:} Compute the largest $m$ singular values and associated singular vectors of $\mathbf{X} = U \Lambda V^{*}$. 
   \STATE \textbf{Step 5:} For all $i \in \cal G$,  $\forall s= 1,2,..m$, compute the spectral kernel descriptor $\psi(\mathbf{x}_{i}, t_{s})$in Eq.(\ref{kernel_signature}) using  $t_{s}$ and the largest $m$ singular vectors and values corresponding to $U $and $\Lambda$, respectively.
   
\end{algorithmic}
\end{algorithm}


\section{Experimental Results}
\label{Experimental_Results}

We evaluate our method on real-world networks in several applications including material science and network alignment of Protein-Protein interactions (PPIs) networks using a recent dataset that has been used to study Covid-19, the SARS-CoV-2. We compare GSE to known and existing state of the art methods in the respective applications. There are a number of benchmarks to evaluate network embeddings built using at least partial supervision, for instance for node classification relative to a given ground truth. Our method does not require any, and therefore we focus our attention on unsupervised benchmarks. Extension of our method to graph classification, or other graph analysis tasks that can exploit supervision is beyond the scope of this paper and will be considered in future work. \\
\textbf{Evaluation metrics:} In the problem of network alignment we use the percentage of correct node correspondence found to evaluate our method, based on the known ground truth correspondence between the two networks. In detecting failure edges in material science we measure the success rate using sensitivity. 
To detect failured edges, we use the graph embedding obtained by each method to cluster the data into two approximately equal size clusters using spectral clustering .
We then choose the cluster with the largest betweenness centrality mean value.  
For further validation, we test the statistical significance of each method by computing the $p$ values using hyper-geometric distribution (see the Appendix). \\
\textbf{Comparison with other methods:}  Our baseline for comparison with the proposed Graph Sylvester Embedding is using the concatenated spectral embedding corresponding to both the Laplacian and the affinity matrix.  We coin the concatenated bases as  St.($\mathbf{W}^{BE},\mathbf{L}^{BE}$). We compare to the spectral Wave Kernel Descriptor signature (WKS) \cite{AubrySC11}  (coined $\mathbf{L}$ desc.). 
$(-)$ in the Tables indicates methods that failed to converge or provided meaningless results.  
\\
We compare with a representative of graph embeddings methods including Laplacian Eigenmaps \cite{Belkin:2003}, node2vec \cite{grover2016node2vec},  and NetMF \cite{UnifyingDeepWalk}. 
We also test against methods that were specifically tailored to this applications we explored including methods for graph alignments \cite{Final,iNeat} and methods based on betweenness centrality for detecting failured edges \cite{Berthier_2019}. 

\subsection{Detect Failed Edges in Material Sciences }
\begin{figure}
\centering
\subfigure[]{
\includegraphics[width=.2\columnwidth]{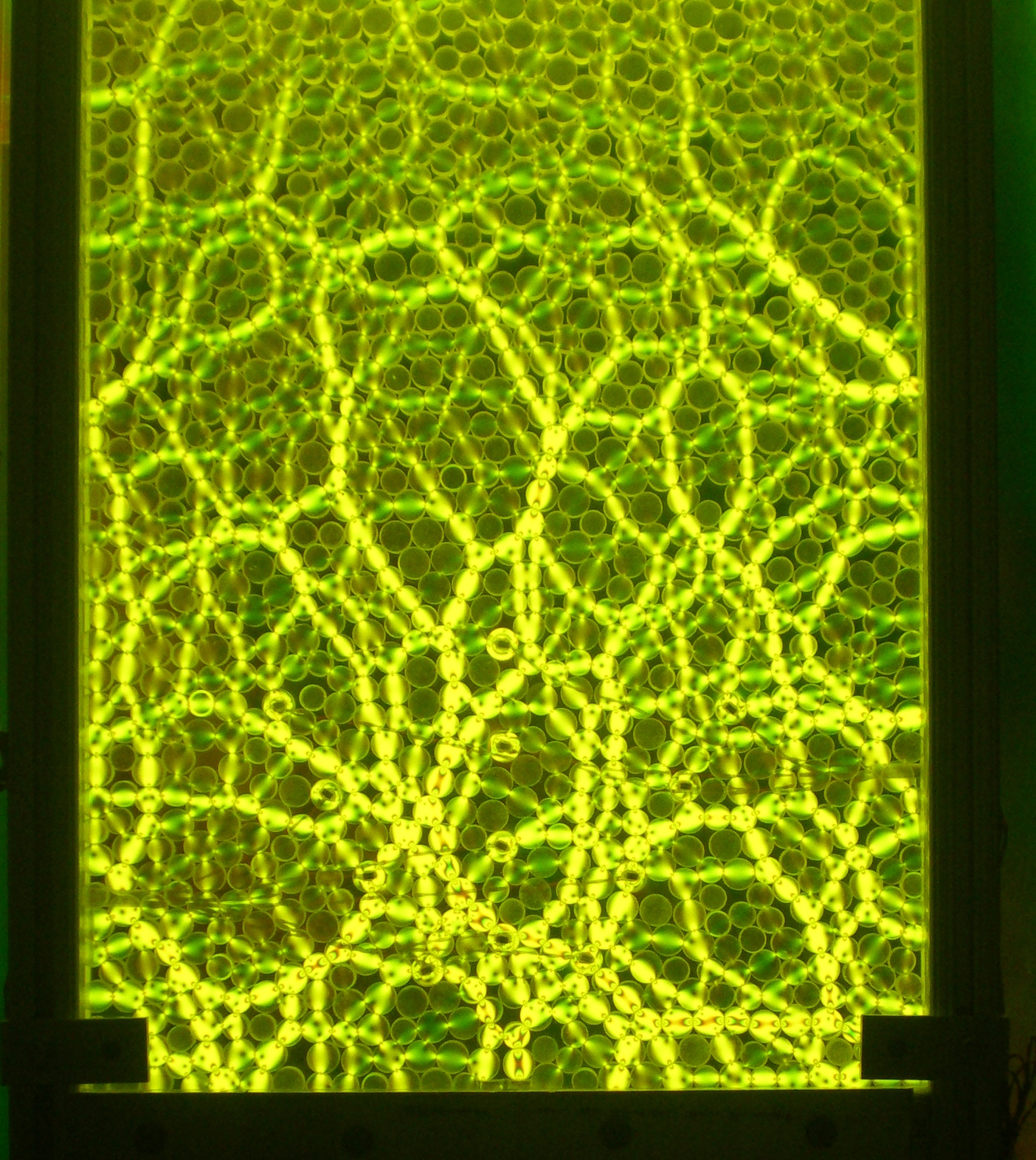}
}
\subfigure[]{
\includegraphics[width=.35\columnwidth]{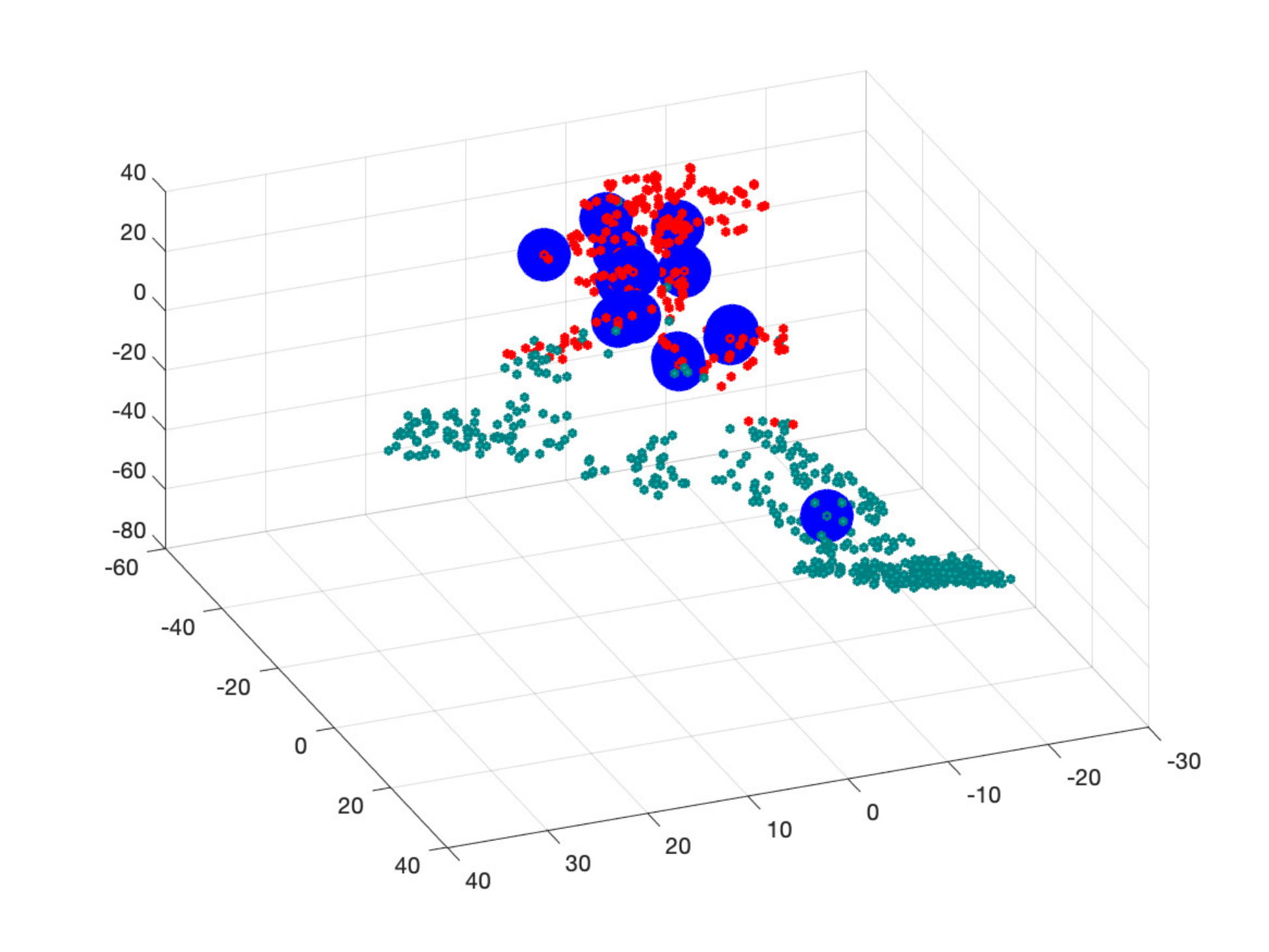}
}
\caption{Graph embeddings used for forecasting failed edges using t-SNE embedding (a) Contact network (yellow), which is extracted from the force chains recorded in a 2D assembly of frictional photoelastic disks overlaid on the reconstructed “pseudo-image" \cite{Berthier_2019}. (b) shows t-SNE visualization of edge embedding using the proposed Sylvester embedding. Points with green color correspond to edges whose value is below the mean, and points with red color correspond to edges whose value is above the mean. The blue enlarged dots correspond to the failed edges in the system which were successfully detected by each method.}
  \label{Granualr_Embedding}
\end{figure}

Forecasting fracture locations in a progressively failing disordered structure is a crucial task in structural materials \cite{Berthier_2019}. Recently, networks were used to represent 2 dimensional (2D) disordered lattices and have been shown a promising ability to detect failures locations inside 2D disordered lattices. 
Due to the ability of BC edges that are above the mean to predict failed behaviors \cite{Berthier_2019}, we expect the edge embedding based on EBC to be effective. As shown in the experimental results below, our proposed \textit{edge} embedding improves robustness in comparison to simpler methods such as the one employed in \cite{Berthier_2019} or methods that are based on "standard" Laplacian embedding. We first describe how to employ edge embedding (proposed node embedding detailed in Sect. \ref{sec:Embedding_Sylvester}) for this task. \\
 \textbf{Edge Embedding using GSE:} Edge embedding is constructed by first applying our node embedding and then using the concatenated nodes features to construct the \textit{edge} embeddings. To forecast the failed edges, we construct a new graph where nodes correspond to the edge embedding features and then apply spectral clustering to cluster the edges into two groups. We apply the same strategy to test spectral descriptors which are based on the graph Laplacian. \\
\textbf{Dataset:} The set of disordered structures was derived from experimentally determined force networks in granular materials \cite{PhysRevMaterialsBerthier}. The network data is available in the Dryad repository \cite{PhysRevMaterialsBerthierDryad}. 
We tested 6 different initial networks, with mean degrees $z ={2.40, 2.55, 2.60, 3.35}$, 3.0, 3.6, following the same datasets corresponding to different initial granular configurations. \\
\textbf{Implementation Details:} In all experiments, we used $c_{t} = 1$ , for all $t$, $c_{t}$ corresponds to the coefficients in the wave kernel signature WKS (Eq (15)). We used a fixed number of total 800 scales $t_{s}$ in all experiments. We measure the success rate in detecting failure edges in terms of the sensitivity (true positive rate).  We report experimental results for a varying number of singular eigenvectors $\mathbf{u}_{l}$ in the SVD of the solution $\mathbf{X}$ to the Sylvester equation, which is then employed to compute edge embeddings (as described above, based on the node embeddings detailed in Algorithm 1 ). \\
Table \ref{tab:Gran_network_with_different_eig} shows a comparison of our method where we report the results using $m= 800$ for all methods which are based on eigensystem computation (this the same number of eigenvectors and associated eigenvalues computed for the spectral methods which are based on graph Laplacian such as LE and spectral descriptors). Fig.\ref{Granualr_Embedding} (b) show visualization using t-SNE of our proposed methods for edge embedding. The points colored in red correspond to the edges whose values are above the mean EBC, while the points colored in turquoise correspond to edges bellow the mean EBC. The blue enlarged dots correspond to the failed edges in the system which were successfully detected by each method. In the examples illustrated, our proposed embeddeding successfully forecast all failed edges, which were mapped to the same cluster including edges whose EBC value was below the mean value.

\begin{table}
  \caption{Success rate ($\%$) in detecting failed edges in granular material networks, comparing GSE to different methods. The network number associated with each Network in parenthesis (Mean deg.) corresponds to the characteristic of the network, given by its mean degree of edges per node. The column below Avg.+ std shows the average and standard deviation summarized over all networks.}
\label{tab:Gran_network_with_different_eig}
  \centering
  \begin{tabular}{lllllllll}
    \toprule
    \cmidrule(r){1-2}
  Method  /Network    & Mean deg.   & Mean deg. & Mean deg. & Mean deg. & Mean deg. & Mean deg. & Avg. $\pm$ std \\
   &  (2.4) &  (2.6)  &    (3.35)  &    (2.55)  &   (3) &   (3.6)  \\
    \midrule
    \midrule
     FL\cite{Berthier_2019}   &  85.7$\%$ & 70$\%$ & 60$\%$&  58.3$\%$  &48.4$\%$ & 58.1$\%$ & 63.4$\%$ $\pm$ 11.7    \\
  NetMF   & 28 $\%$  &  40 $\%$ &  72 $\%$  & 45 $\%$  & \textbf{93.9 }$\%$ &  90.0 $\%$ & 61.4$\%$ $\pm$  25.2 \\   
    node2vec  &  64.2$\%$   & 55$\%$   & 40$\%$  & \textbf{79.0} $\%$& 69.6$\%$ & 37.2$\%$ & 61.4$\%$ $\pm$ 15.1 \\
     LE   &  64.2$\%$   & 75$\%$& 68.0$\%$ &  66.6$\%$ &  63.6$\%$  &  \textbf{83.7$\%$} & 70.1$\%$ $\pm$ 6.5 & \\
  $\mathbf{L}$ desc.  & 57.1$\%$ & 45$\%$ & \textbf{84.0$\%$}  & 45.0$\%$ & 54.5$\%$ & 37.2$\%$&  53.8$\%$ $ \pm$  15 &\\
   St.($\mathbf{W}^{BE},\mathbf{L}^{BE}$) & 64.2$\%$  & 75.0$\%$ & 80.0$\%$ & 75.0$\%$ & 78.7$\%$ & 79.0$\%$ & 75.3$\%$ $\pm$ 27.6 &\\
    \textbf{GSE}  & \textbf{92.8}$\%$ & \textbf{80}$\%$ & 76.0$\%$ &  75.0$\%$  & 72.2$\%$ & 83.7$\%$ & \textbf{79.9 $\%$}$\pm$  6.8 & \\
    \bottomrule
  \end{tabular}
\end{table}

\subsection{Network Alignment on PPI networks with human protein interactors of SARS-CoV-2}
Effective representation of nodes with similar network topology is important for network alignment applications, where the network structure around each node provides valuable information for matching and aligning the networks. Network alignment of Protein-protein interactions (PPIs) networks is considered as an important first step for the analysis of biological processes in computational biology. Typically, popular methods in this domain gather statistics about each node in the graph (e.g: node degree) followed by an optimization methods to align the network. For PPI networks, it is assumed that a protein in one PPI network is a good match for a protein in another network if their respective neighborhood topologies are a good match \cite{Isorank}. In this case, features or nodes attributes are in general not available which makes this task of network alignment and finding node correspondence more challenging, with only the graph network to rely on. \\

\begin{figure}
\centering
\subfigure[]{
\includegraphics[width=.25\columnwidth]{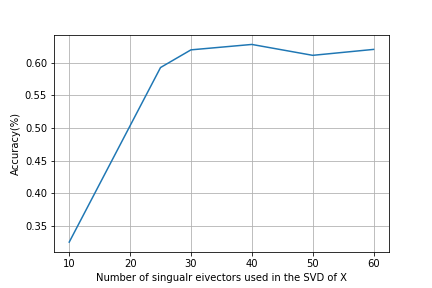}
}
\subfigure[]{
\includegraphics[width=.25\columnwidth]{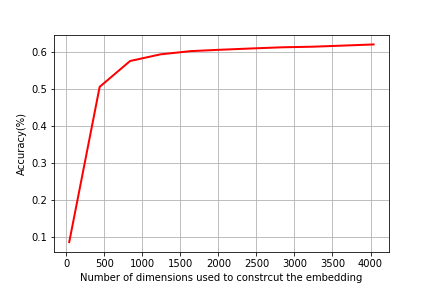}
}
\caption{Sensitivity to parameter selection in network alignment of the STRING network, concerning the number of singular eigenvectors and the number of scales used in the spectral signature of GSE (a) Accuracy (in percentage of nodes whose nearest neighbor corresponds to its true node correspondence) as a function of the number of singular eigenvectors used in the SVD of $\mathbf{X}$  in the network alignment of STRING. (b) Shows the accuracy as a function of the number of scales which was used to construct the Spectral Kernel descriptor for the network alignment of the STRING network.}
  \label{Abalation_Study1}
\end{figure}
\textbf{Dataset:} We test using the STRING network from the STRING database, a PPI network that consists of 18,886 nodes and 977,789 edges. The STRING network includes 332 human proteins that physically interact with SARS-CoV-2. It is likely to have both false positives and false negatives edges. We aim to find network alignment between two copies of the STRING network, using an additional STRING network which is created by randomly removing 10$ \%$ of the network edges. We also test on the Lung network \cite{Lung}, which is a more recent PPI networks with known human protein interactors of SARS-CoV-2 proteins that consists of 8376 nodes, 48522 edges, and 252 human proteins that physically interact with SARS-CoV-2. 
\textbf{Implementation details:} Given the node correspondence of the protein interactors of SARS-CoV-2, we connect each pair of nodes corresponding to the same human protein interactors of SARS-CoV-2 with an edge, which is resulted in a network composed of the two STRING networks. We then apply the proposed GSE  to generate node embeddings. We used 5-fold cross validation, where in each experiment we used 50$\%$ of  the known human proteins in the STRING and Lung networks. \\
We compare to network alignment methods, such as Isorank \cite{Isorank}, Final \cite{Final} and iNEAT \cite{iNeat}. For the network alignments methods, we provide as an input the corresponding affinity graphs and a matrix with the known correspondence between the nodes that correspond to human proteins that physically interact with SARS-CoV-2. Additionally, we also tested graph embeddings including Laplacian Eigenmaps (LE) \cite{Belkin03}, Locally Linear Embedding (LLE) \cite{LLE}, Hope \cite{Hope}, and node2vec \cite{grover2016node2vec} which are given the same input network. Graph embedding methods are not effective for this task and thus some of the results are omitted. Network alignments methods perform better, while our proposed GSE outperforms all competing methods.\\
Table \ref{tab:Network_Align_Lung_String} shows the percentage of nodes whose nearest neighbor corresponds to its true node correspondence using GSE in the STRING and Lung networks. Our proposed GSE outperforms all competing methods.  \\

\begin{table}[t]
\caption{Classification accuracy on network alignment using PPI interaction networks: (STRING(s) and Lung.}
\label{sample-table}
\vskip 0.15in
\begin{center}
\begin{small}
\begin{sc}
\begin{tabular}{lcccr}
\toprule
Method/Data & STRING (s)  & Lung &   \\
\midrule
node2vec   &   10.4 \%  & - \% &\\
Isorank    &    23 \%  & 44.2 \% &\\
Isorank using $\BCG$   & 20 \%     & 44.2 \%   \\
Final   &  36 \%   &  50.5   \%  & \\
iNeat  & 37.0  \% &  53.5 \% &  \\
 St.($\mathbf{W}^{BE},\mathbf{L}^{BE}$) & 25.3  \% &  53.2 \%   &  \\
GSE   &  \textbf{61.2$\%$}   &  \textbf{55.0 \%}  & \\
\bottomrule
\end{tabular}
\end{sc}
\end{small}
\end{center}
\vskip -0.1in
  \label{tab:Network_Align_Lung_String}
\end{table}


\section{Discussion}
\label{sec: Discussion}

We have presented an embedding to represent local and global structure of a graph network, constructed without the need for any supervision nor explicit annotation. Besides the cost of time and effort, annotating data can create privacy and security risks. Our focus is on developing expressive and flexible representations that can be used in a variety of downstream tasks without human intervention. Flexibility is exercised through the choice of bases, that are combined through Sylvester's equation. In the specific cases we have experimented, the bases are chosen to capture local and structural similarity across the graph network. 




The implementation of our approach is computationally intensive, with most of the burden falling on the computation of the BCG, which is $O(mN)$ where $m$ is the total number of edges and $N$ is the total number of nodes, thus approximately $O(N^2) $ for sparse graphs. The execution time of our method with Python code implementation using Intel Core i7 7700 Quad-Core 3.6GHz with 64B Memory on the Cora dataset with approximately 2700 nodes takes $\approx$  24.9 seconds. 
 
 
Computing BC measurements on large graphs is computationally heavy ($O(N^2) $ for sparse graphs), and developing fast methods for BC measurements is an ongoing research area. With significant progress that has been made in recent years (e.g: \cite{FastBC}) we can foresee extensions to scale our approach to networks with millions of nodes. In addition, the effectiveness of BC measurements degrades on larger graphs due to noise (for example when the graph includes a large number of nodes with clustering coefficient close to zero). Applying our approach to small sub-graphs or ``Network motifs" before aggregating it to to the entire graph may be one foreseeable solution.



In our study, we found that spectral embedding by solving the Sylvester equation of the edge betweeness centrality graph reveals the node's network structure, which is not necessarily possible with the original graph weights. It is also possible to compute edge betweenness centrality in a way that takes into account the original weight information. For example, by looking into the path evaluation function that assesses a path between two nodes that is combining both the sum of edge weights and the number of shortest distance path as was proposed in \cite{Path_Evaluation_Centralitie}. Another possible direction is to solve a generalized Sylvester equation which would incorporate both matrices corresponding to the original weight information and the edge betweenness centrality. 
Other future work includes expanding our approach to address applications which include dynamic graphs and multi-layer graphs, which would include solving Sylvester equation with time-varying coefficients. 


\nocite{langley00}



\bibliographystyle{plain}
\bibliography{egbib.bib}



\section{Appendix}

 \subsection{Graph Differentiation  }
The lower boundary of the curve enclosing the feasibility domain of the spreads $\left (\mathbf{s}_{\mathbf{L}^{BE} }(\mathbf{x}) , \mathbf{s}_{\mathbf{W}^{BE} }(\mathbf{x})  \right )$, with respect to a unit norm vector $\mathbf{x} \in l^{2}(\cal G) $ is defined as 
\begin{equation}
\label{Feasiblility_Domain}
\left\{\begin{matrix}
\Gamma_{ \mathbf{s}_{\mathbf{W}^{BE}  } } (\mathbf{s}_{\mathbf{L}^{BE} }) =  \underset{\mathbf{x}}{\mbox{min}} \, \,
 g_{  \mathbf{W}^{BE} }   (\mathbf{x})  & \\ 
\mbox{s.t} \,  \,   g_{  \mathbf{L}^{BE} }(\mathbf{x})  =  \mathbf{s}_{\mathbf{L}^{BE} }  \, \mbox{and}  \, \, \mathbf{x}^{T}\mathbf{x} =1 & 
\end{matrix}\right.
\end{equation}
To solve (\ref{Feasiblility_Domain}), we use the Lagrangian 
\begin{equation}
L(\mathbf{x}, \beta , \gamma)=  \, \,\mathbf{x}^{T}  \mathbf{W}^{BE} \mathbf{x} - \beta ( \mathbf{x}^{T}\mathbf{L}^{BE}\mathbf{x} ) - \gamma (\mathbf{x}^{T}\mathbf{x}  - 1) 
\end{equation}with $\beta \in \mathbb{R}$.  
Differentiating 
and comparing 
to zero, we obtain the following eigenvalue problem: 
\begin{equation}
\label{generalized eigenvalue problem}
( \mathbf{W}^{BE}   - \beta \mathbf{L}^{BE}) \mathbf{x} =  \gamma \mathbf{x}  
\end{equation}
where the eigenvector $\mathbf{x}$ solving (\ref{generalized eigenvalue problem}) is also a minimizer for (\ref{Feasiblility_Domain}). 
Denoting  
\begin{equation}
\label{Generlaized_Lap}
\mathbf{\tilde{L}}(\beta) =  \mathbf{W}^{BE}   -  \beta\mathbf{{L}}^{BE} 
\end{equation}we can write the generalized eigenvalue problem (\ref{generalized eigenvalue problem}) using the matrix pencil $\mathbf{\tilde{L}}(\beta)$ 
\begin{equation}
\mathbf{\tilde{L}}(\beta)\mathbf{x} =  \gamma \mathbf{x}  
\end{equation}
The scalar $\beta \in \mathbb{R}$  controls the trade-off between the total vertex and spectral spreads, as illustrated in the barbell graph $\tilde{\mathbf{L}}^{BE}(\beta$) in Fig.\ref{Eigenvectors_Barbell} (a) and (b) using $\beta$ = -200, and $\beta$ = -0.2, respectively. Functions colored with green correspond to the eigenvectors of $\tilde{\mathbf{L}}^{BE}(\beta$),  functions colored with blue correspond to the eigenvectors of ${\mathbf{L}}^{BE}$.
As can be seen, when  $|\beta|$ is large,  the eigenvectors of $\tilde{\mathbf{L}}^{BE}(\beta$)  reveal structure which is similar to the eigenvectors of ${\mathbf{L}}^{BE}$, while small values of $|\beta|$ produce structure which is similar to those corresponding to ${\mathbf{W}}^{BE}$.\\  
The analysis above yields the generalized graph Laplacian in Eq.(\ref{generalized eigenvalue problem}) whose solutions can be used for graph embedding. One could use different scaling coefficients for the two matrix coefficients, thus giving different weight to the Laplacian and affinity terms in the resulting embedding. The embedding (coined GSSE), obtained by solving Eq.(\ref{Generlaized_Lap}), is composed of solutions to the generalized eigenvalue problem that characterize the relationships between these two quantities, specifically the lower bound of their feasibility domain.

\subsection{Node Embeddings using Total Vertex and Spectral Spreads (GSSE)}
\label{Sec:GSSE}
We present an additional embedding method which is based on the analysis in section 4.2 (Graph Differentiation) which derived the generalized Laplacian $\mathbf{\tilde{L}}$ in Eq.(10). 
In order to employ  $\mathbf{\tilde{L}}$ for analysis and practical considerations it is often useful to encode the network captured in $\mathbf{\tilde{L}}$ using a semi positive definite operator. 
For practical consideration, we transform $\mathbf{\tilde{L}}$ into a semi positive definite matrix $\mathbf{L}_{\Delta}$ using a simple perturbation matrix $\Delta$, where $\Delta = \mu \mathbf{I} $, $\mu = - \tilde{\lambda}_{0} $, where $\tilde{\lambda}_{0} $ is the smallest eigenvalue of $\mathbf{\tilde{L}}$, and $\mathbf{I}
$ is the identity matrix. 
Setting 
 \begin{equation}
 \label{SPD_addition}
 \mathbf{L}_{\Delta} =   \mathbf{\tilde{L}} + \Delta 
 \end{equation}ensures that $\mathbf{L}_{\Delta}$ is semi positive definite (SPD) matrix. Letting $\left \{ \tilde{\lambda}_{i} \right \}_{i=1}^{N}$ and $\left \{ \lambda^{\Delta}_{i} \right \}_{i=1}^{N}$ be the eigenvalues corresponding to $\mathbf{\tilde{L}} $ and  $\mathbf{{L}}_{\Delta}$, respectively, we can see that the choice made in (\ref{SPD_addition}) ensures that the original spacing in eigenvalues of $ \mathbf{\tilde{L}} $ is preserved in $\mathbf{{L}}_{\Delta}$ $
 \lambda^{\Delta}_{i+1} - \lambda^{\Delta}_{i} = \tilde{\lambda}_{i+1} -  \tilde{\lambda}_{i}$.  The embedding method proposed using $ \mathbf{L}_{\Delta} $ is coined Graph Spectral Spread Embedding (GSSE). Note that there is a geometric interpretation which is related to the way $\mathbf{\tilde{L}} (\beta)$ was obtained from the lower boundary curve enclosing the feasibility domain  $\mathbb{D}_{ ( s_{\mathbf{L} },  s_{\mathbf{W}}  )  } $;  Setting instead another perturbation matrix $\Delta = \mu \mathbf{I} $  with $\mu = - \lambda_{N}^{W} $ and defining
 
\begin{equation}
q(\beta)  ={ \mbox{min}  } ( {\lambda } (\mathbf{L}_{\Delta} (\beta) ) )
\end{equation}
where ${ \mbox{min}  } ( {\lambda } (\mathbf{L}_{\Delta} (\beta) ) ) $ corresponds to the minimum eigenvalue of $\mathbf{L}_{\Delta} $ .
we have that 
\begin{equation}
 \mathbf{g}_{\mathbf{W} }(\mathbf{x})- \beta \mathbf{g}_{\mathbf{L}} (\mathbf{x})  \geq q(\beta)
\end{equation}which defines a half plane in $\mathbb{D}_{ ( s_{\mathbf{L} },  s_{\mathbf{W}}  )  }$. 

\textbf{Properties:} The feasibility domain $\mathbb{D}_{ (s_{\mathbf{L}^{BE} }, s_{\mathbf{W}^{BE} })  } $ is a bounded set since  $\forall  (s_{\mathbf{L}^{BE}}, s_{\mathbf{W}^{BE} }) \in \mathbb{D}_{ (s_{\mathbf{L}^{BE}}, s_{\mathbf{W}^{BE} })   } $ 
\begin{equation}
0  \leq s_{\mathbf{L}^{BE} } \leq  \lambda_{N}^{BE}  \, \, \mbox{and}\\
- \lambda_{N}^{W^{BE}} \leq s_{\mathbf{W}^{BE}}  \leq  \lambda_{N}^{W^{BE}}
\end{equation}
where $\lambda_{N}^{W^{BE}}$corresponds to the largest eigenvalue of the affinity graph.

\subsection{Forecasting failures edges: additional details and  comparisons}
\label{submission}
We report additional experimental results on the granular material datasets  \cite{Berthier_2019}. In all experiments, we used $c_{t} = 1$ , for all $t$, $c_{t}$ corresponds to the coefficients in the wave kernel signature WKS (Eq.(\ref{kernel_signature})). We used a fixed number of 800 scales $t_{s}$ and a fixed and number of singular eigenvectors corresponding to the number of points. The experiments reported in Table \ref{tab:Gran_network_p_values}) test the statistical significant of each method by computing the $p$ values using hyper-geometric probability distribution. Specifically,  the total number of edges corresponds to the total population size parameter in the hyper-geometric distribution with the feature that contains K failed edges, the size of the cluster is the number of draws and the number of observed successes corresponds to the number of edges correctly classified as failed edges.  

\begin{table}
  \caption{Additional validation on the granular material datasets represented by irregular networks: we report $p$ values, testing the statistical significance of our model for detecting failed edges. The $p$ values are computed using the hypergeometric distribution.}
\label{tab:Gran_network_p_values}
  \centering
  \begin{tabular}{llllllll}
    \toprule
    \cmidrule(r){1-2}
  Method  /Network    & Mean deg.   & Mean deg. & Mean deg. & Mean deg. & Mean deg. & Mean deg. & \\
   &  (2.4) &  (2.6)  &    (3.35)  &    (2.55)  &   (3) &   (3.6)  \\
    \midrule
    \midrule
    node2vec  & $ 1.9 \cdot  10^{-1}  $ $10^{-1}$ &  3 $\cdot 10^{-2}$  &  $ 3.4\cdot 10^{-2}$ & $  2\cdot 10^{-1}$ &   $ 6.3 \cdot 10^{-1}$ &  $ 5\cdot  10^{-2}$ & \\
  NetMF   & $ 8\cdot  10^{-1}$  & $  8\cdot 10^{-1} $  &  $  7.3\cdot  10^{-3}$ & $  5\cdot 10^{-1}$  &  $ 2.2 \cdot  10^{-9} $ & $ 2.4 \cdot 10^{-9}$  & \\
LE   & $ 7.4\cdot  10^{-1}$  & $ 8.3\cdot  10^{-3} $ & $ 2.5\cdot  10^{-1}$  & $ 5\cdot 10^{-1}$  & $ 5\cdot 10^{-3} $  & $ 2\cdot  10^{-2}$   & \\
  $\mathbf{L}$ desc. & $  4\cdot  10^{-1}$  &  $ 7\cdot  10^{-1}$   & $  2.1 \cdot  10^{-2}$    &  $ 5 \cdot  10^{-1}$ & $ 6.5\cdot  10^{-1}$ & $ 7 \cdot  10^{-1}$ & \\
  $\mathbf{W}^{BE}$ desc. &   $ 3.4 \cdot 10^{-1}$  & $ 6.1\cdot  10^{-1}$  &  $3.8  \cdot  10^{-3} $  & 9.9 $ \cdot 10^{-3}$    & $  0.1\cdot  10^{-1}$  & $ 3.4 \cdot 10^{-1}$  &\\
   St.($\mathbf{W}^{BE},\mathbf{L}^{BE}$) & $ 2.4\cdot 10^{-3}$   & $ 7.1\cdot 10^{-4}$ & $   3\cdot  10^{-2}$  & $  3\cdot  10^{-2}$  &  $ 4\cdot  10^{-2}$  & $  3.2\cdot 10^{-2}$  &\\
   \textbf{GSE}    & $ 2\cdot 10^{-4} $  &  $ 2.6\cdot 10^{-2}$   & $  1\cdot  10^{-1}$ &   $4\cdot 10^{-5}$ &  $ 1.6\cdot 10^{-3}$  & $ 2.9\cdot  10^{-2}$    &\\
    \bottomrule
  \end{tabular}
\end{table}

\section{Network Alignment}

We report additional results and details in the problem of network alignment using the STRING network from the STRING database. In the experimental results we used all 332 human proteins known to interact with the Sars-Cov-2 as the available node correspondence between the two networks we tested on network alignment. From the 18,886 node of the String network we extract the 1000 nodes which corresponds to the nodes with the largest diffusion scores and the 332 nodes corresponding the human proteins known to interact with the Sars-Cov-2, based on the method suggested in
\cite{gigascience}. Alignment is performed between two copies of the STRING network, where 10$\%$ and 20$\%$ edges were removed from the sub-network consisting of a total of 1332 nodes. It is evident that graph embedding methods such as LLE and LE, which are rooted in manifold learning that is biased to local similarity and heavily relies on the graph smoothness are not effective for this task. 

\begin{table}
 \caption{Network Alignment using  the STRING network: percentage of nodes whose nearest neighbor corresponds to its true node correspondence using GSE compared to Graph Embedding methods (using all 332 available correspondence).}
\label{tab:Noisy_STRING_dataset}
  \centering
  \begin{tabular}{lll}
    \toprule
    \cmidrule(r){1-2}
Method/noisy edges percentage & 10$\%$ & 20$\%$  \\
    \midrule
 LE \cite{Belkin03} &  5.38$\%$ & -     \\  
 LLE \cite{LLE}  & 1.5$\%$  & 1.3$\%$   \\
RL \cite{Reg_Lap} &  1$\%$ & -  \\
HOPE  \cite {Hope}&  2$\%$ &- \\
 Isorank \cite{Isorank} & 41$\%$ & 40 $\%$ \\
 Final \cite{Final} & 58.2 $\%$  & 56.6$\%$ \\
iNEAT \cite{iNeat}  &  63.8$\%$   & 56.1$\%$ \\
\textbf{GSSE} &  48 $\%$  &  20.21 $\%$  \\
\textbf{GSE} & \textbf{76.4$\%$}   & \textbf{60}$\%$ \\
 \bottomrule
 \end{tabular}
\end{table}

\section{The Sylvester operator}
\label{Comp_Graph_Learning}

The discrete-time Sylvester operator $\mathbf{S}(\mathbf{X}) = \mathbf{A} \mathbf{X}\mathbf{B}  - \mathbf{X} $  is used to express the eigenvalues and eigenvectors of $\mathbf{A}$ and $\mathbf{B}$ using a single operator $\mathbf{S}$, where we solve
\begin{equation}
\label{eqn:Sylvester}
\mathbf{S}(\mathbf{X})   =  \mathbf{C} 
\end{equation}
using $\mathbf{C} =   \mathbf{I} $,  ($\mathbf{I} \in \mathbb{R}^{N\times N}$ is the identity matrix), $\mathbf{A} = \mathbf{W}^{BE}, \mathbf{B}  = \mathbf{L}^{BE}$. 

\begin{Prop}
Suppose that $\mathbf{P} \in \mathbb{R}^{N \times N}$ is a permutation matrix, $\mathbf{\tilde{A}} = \mathbf{P}^{T}\mathbf{A} \mathbf{P}$,  $\mathbf{\tilde{B}} = \mathbf{P}^{T}\mathbf{B} \mathbf{P}$, and 
assume that $\mathbf{X}, \mathbf{\tilde{X} } $ are two solutions to the discrete-time Sylvester equation Eq.(\ref{eqn:Sylvester})  with the associate matrices  $\mathbf{{A}}, \mathbf{{B}}$ and  $\mathbf{\tilde{A}} ,   \mathbf{\tilde{B}}$,  respectively, then 
\begin{equation}
\mathbf{\tilde{X}} = \mathbf{P}^{T}\mathbf{X} \mathbf{P}
\end{equation}
\end{Prop}
\textbf{Proof:} For a permutation matrix $\mathbf{P} \in \mathbb{R}^{N \times N}$, let 
$\mathbf{\tilde{A}} = \mathbf{P}^{T}\mathbf{A} \mathbf{P}$, and  $\mathbf{\tilde{B}} = \mathbf{P}^{T}\mathbf{B} \mathbf{P}$.
Let $\mathbf{X}, \mathbf{\tilde{X} } $ are two solutions to the discrete-time Sylvester equation Eq.(\ref{eqn:Sylvester}) with the associate matrices $\mathbf{{A}}, \mathbf{{B}}$ and  $\mathbf{\tilde{A}} ,   \mathbf{\tilde{B}}$,  respectively,   $\mathbf{A} \mathbf{X}\mathbf{B}  - \mathbf{X}  =  \mathbf{I} $ and  $\mathbf{\tilde{A}} \mathbf{\tilde{X}}\mathbf{\tilde{B}}  - \mathbf{\tilde{X}}  =  \mathbf{I} $. 
Next note that $ \mathbf{P}^{T}\mathbf{A} \mathbf{P} \mathbf{\tilde{X}}\mathbf{P}^{T}\mathbf{B} \mathbf{P}   - \mathbf{\tilde{X}}  =  \mathbf{I} $.  Since $\mathbf{P}$ is a permutation matrix, we have that $\mathbf{P}^{-1} = \mathbf{P}^{T} $  and by multiplying  $\mathbf{P}$ and $\mathbf{P}^{T}$ from the left hand and right hand side, respectively, we obtain  $\mathbf{A} \mathbf{P} \mathbf{\tilde{X}}\mathbf{P}^{T}\mathbf{B}   - \mathbf{\tilde{X}}  =  \mathbf{I} $.  Denoting $ \mathbf{Y} = \mathbf{P} \mathbf{\tilde{X}}\mathbf{P}^{T}$,  under the assumptions that $\mathbf{{X}}$ is a unique solution, then $\mathbf{{X}} = \mathbf{Y} = \mathbf{P} \mathbf{\tilde{X}}\mathbf{P}^{T}$.$\square$ \\

The resulting solution of $\mathbf{X}$ is described in the next Lemma below.  
Note that since $\mathbf{A} = \mathbf{W}^{BE}, \mathbf{B}  = \mathbf{L}^{BE}$ we have $\mathbf{A}$ and $\mathbf{B}$ that are diagonalizable,  (with the matrices $\Phi^{\mathbf{W}^{BE}}, \Phi^{\mathbf{L}^{BE}}$ corresponding to the associated eigenvectors of  $\mathbf{W}^{BE}, \mathbf{L}^{BE}$, respectively). 

\begin{Lemma}
\label{Eig_Sylvester}
Let $\mathbf{A}, \mathbf{B}$, and $ \mathbf{C}  \in \mathbb{R}^{N\times N} $ in the Sylvester equation  (\ref{eqn:Sylvester}), with 
$\mathbf{A} = \mathbf{W}^{BE}, \mathbf{B}  = \mathbf{L}^{BE}$. Let $   \left \{u_{l}  \right \}_{l=1}^{N} $, $ \left \{  \lambda_{l}   \right \} _{l=1}^{N}$, and 
$  \left\{  v_{l}  \right \}_{l=1}^{N} $, $ \left \{  \mu_{l}   \right \} _{l=1}^{N}$ be the corresponding eigenvectors and eigenvalues of $\mathbf{A}$ and $  \mathbf{B}$, respectively.  Assume that $ \lambda_{i}^{\mathbf{W}^{BE}}\lambda_{j}^{\mathbf{L}^{BE}}  \neq 1 \, \forall i, j $,  and that the graph representation encoded in $\mathbf{A}$ and $\mathbf{B}$ have the same order based on the identity map , $\pi : \mathbf{A}\rightarrow \mathbf{B}$ with $\pi( \mathbf{A}(:,i)) =\mathbf{B}(:,i)$ for each $i \in \cal V$. Then, if $ \mathbf{C}=  \mathbf{I}$, the solution $\mathbf{X}$  to the discrete-time Sylvester system (\ref{eqn:Sylvester}) is
\begin{equation}
\mathbf{X} = \Phi^{\mathbf{W}^{BE}}\mathbf{\tilde{C}} (\Phi^{\mathbf{L}^{BE}})^{T}
\end{equation}
where $\mathbf{\tilde{C}}_{i,j} = \frac{   (\phi_{i}^{\mathbf{W}^{BE}})^T \phi_{j}^{\mathbf{L}^{BE}}    }   {\lambda_{i}^{\mathbf{W}^{BE}}\lambda_{j}^{\mathbf{L}^{BE}} - 1} $
\end{Lemma}
\textbf{Proof:} Using $\mathbf{A} = \mathbf{W}^{BE}, \mathbf{B}  = \mathbf{L}^{BE}$ we have that $\mathbf{A}$ and $\mathbf{B}$ are diagonalizable, 
$\mathbf{A}  = \Phi^{\mathbf{W}^{BE}} \Lambda  (\Phi^{\mathbf{W}^{BE}}) ^{-1}$ and $\mathbf{B}  = \Phi^{\mathbf{L}^{BE}} \Lambda  (\Phi^{\mathbf{W}^{BE}}) ^{-1}$ where 
$\Lambda^{\mathbf{W}^{BE} } =  \mbox{diag} ( \lambda_{1}^{\mathbf{W}^{BE} } ,... ,\lambda_{N}^{\mathbf{W}^{BE} }   )$  and $\Lambda^{\mathbf{L}^{BE} } =  \mbox{diag} ( \lambda_{1}^{\mathbf{L}^{BE} } ,... ,\lambda_{N}^{\mathbf{L}^{BE} }   )$. Since the matrices $\Phi^{\mathbf{W}^{BE}} , \Phi^{\mathbf{L}^{BE}}$ are orthogonal, we have 
\begin{equation}
\Phi^{\mathbf{W}^{BE}}\Lambda^{\mathbf{W}^{BE} } (\Phi^{\mathbf{W}^{BE}})^{T} \mathbf{X} \Phi^{\mathbf{L}^{BE}}\Lambda^{\mathbf{L}^{BE} } (\Phi^{\mathbf{L}^{BE}})^{T} - \mathbf{X} = \mathbf{C}
\end{equation}

Multiplying  by $(\Phi^{\mathbf{W}^{BE} })^{T} $ and $\Phi^{\mathbf{L}^{BE} } $ we obtain 
\begin{equation}
\Lambda^{\mathbf{W}^{BE} } (\Phi^{\mathbf{W}^{BE}})^{T} \mathbf{X} \Phi^{\mathbf{L}^{BE}}\Lambda^{\mathbf{L}^{BE} }- (\Phi^{\mathbf{W}^{BE}})^{T}\mathbf{X} \Phi^{\mathbf{L}^{BE} } = (\Phi^{\mathbf{W}^{BE} })^{T} \mathbf{C} (\Phi^{\mathbf{L}^{BE} })
\end{equation}
  
Setting $ \mathbf{\tilde{C}} = (\Phi^{\mathbf{W}^{BE}})^{T}\mathbf{X} \Phi^{\mathbf{L}^{BE} } $ and using $ \mathbf{C}=  \mathbf{I}$, we have 

\begin{equation}
\Lambda^{\mathbf{W}^{BE} }  \mathbf{\tilde{C}} \Lambda^{\mathbf{L}^{BE} }-  \mathbf{\tilde{C}}= (\Phi^{\mathbf{W}^{BE} })^{t} (\Phi^{\mathbf{L}^{BE} })
\end{equation}
Since $\Lambda^{\mathbf{W}^{BE}}$ is a diagonal matrix, we have that the $i$th row of $\Lambda^{\mathbf{W}^{BE} }  \mathbf{\tilde{C}}\Lambda^{\mathbf{L}^{BE} }$ is $\lambda_{i}$ times the $i$th row of $ \mathbf{\tilde{C}}\Lambda^{\mathbf{L}^{BE} }$and since $\Lambda^{\mathbf{L}^{BE} }$ is a diagonal matrix, then the $i$th column of $\Lambda^{\mathbf{W}^{BE} }  \mathbf{\tilde{C}} \Lambda^{\mathbf{L}^{BE} }$ is $\lambda_{i}^{\mathbf{L}^{BE} } $ times the $i$th column of $\Lambda^{\mathbf{W}^{BE} } \mathbf{\tilde{C}}$. Combining the two properties, we obtain that for the $(i,j)$ entry $$\lambda_{j}^{\mathbf{L}^{BE} } \lambda_{i}^{\mathbf{W}^{BE} }\tilde{c}_{i,j} - \tilde{c}_{i,j}  =  (\phi_{i}^{\mathbf{W}^{BE}})^T \phi_{j}^{\mathbf{L}^{BE}}$$ hence: 
 $\mathbf{\tilde{C}}_{i,j} = \frac{   (\phi_{i}^{\mathbf{W}^{BE}})^T \phi_{i}^{\mathbf{L}^{BE}}    }   {\lambda_{i}^{\mathbf{W}^{BE}}\lambda_{j}^{\mathbf{L}^{BE}} - 1} $, with $\mathbf{X} = \Phi^{\mathbf{W}^{BE}}\mathbf{\tilde{C}} (\Phi^{\mathbf{L}^{BE}})^{T}$.$\square$ \\
The next Proposition is a result of  Lemma 2.1 in \cite{Poly_Sol_Sylvester}, showing  the solution $\mathbf{X}$ to the Sylvester equation as a polynomial of the matrices $\mathbf{A}$ and  $\mathbf{B}$ \cite{Poly_Sol_Sylvester}, where in our case using higher-order polynomials in $\mathbf{W}^{BE}$, $\mathbf{L}^{BE}$ can be interpreted as imposing smoothness in the embedding space.

\begin{Prop} 
Given $\mathbf{A}, \mathbf{B}$, and $ \mathbf{C}  \in \mathbb{R}^{N\times N}$, where $ \mathbf{A}, \mathbf{B}$ do not share eigenvalues (i.e., $\sigma(\mathbf{A}) \cap  \sigma(\mathbf{B})  =\emptyset $). Let   $ \eta (k,\mathbf{A}, \mathbf{B},\mathbf{C}) =  {\mathbf{A}}^{k}\mathbf{X}  - \mathbf{X}  {\mathbf{B}}^{k}   $ and $\mathbf{C}=  \mathbf{I}$, then

\begin{equation}
{ \mathbf{A}}^{k}\mathbf{X}  - \mathbf{X}  {\mathbf{B}}^{k} =  \sum_{i=0}^{k-1}  \mathbf{A}^{k-1-i} {\mathbf{B}}^{i}   
\end{equation}Moreover, the solution to the Sylvester equation can be represented as a polynomial in $\mathbf{A}, \mathbf{B}$ for $k\geq 1$,  $d \geq 1$ :
\begin{equation}
\mathbf{X} =  q(\mathbf{A}^{d})^{-1} \eta (k,\mathbf{A}^{d} , \mathbf{B}^{d})
\end{equation} where $\eta (k,\mathbf{A}^{d} , \mathbf{B}^{d})  = { \mathbf{A}}^{kd}\mathbf{X}  - \mathbf{X}  {\mathbf{B}}^{kd}  $ and $q(\mathbf{A}^{d})$ is the characteristic polynomial of $\mathbf{A}^{d}$. 
\end{Prop} 
\textbf{Proof:}
Let $\mathbf{X}$ be a solution to Eq (\ref{eqn:Sylvester}), we assume that $\mathbf{A}  \in \mathbb{R}^{N\times N}$and $\mathbf{B} \in \mathbb{R}^{N\times N}$are symmetric matrices with the same dimensionality, and that $\sigma(\mathbf{A}) \cap  \sigma(\mathbf{B})  =\emptyset $. By Lemma 2.1 in \cite{Poly_Sol_Sylvester} $\mathbf{X}$ is unique and
\begin{equation}
{ \mathbf{A}}^{k}\mathbf{X}  - \mathbf{X}  {\mathbf{B}}^{k} =  \sum_{i=0}^{k-1}  \mathbf{A}^{k-1-i}  \mathbf{C} {\mathbf{B}}^{i}   
\end{equation}for any $k \geq 1$ .
Taking $\mathbf{C} =  \mathbf{I}$, we have that  $  { \mathbf{A}}^{k}\mathbf{X}  - \mathbf{X}  {\mathbf{B}}^{k} =  \sum_{i=0}^{k-1}  \mathbf{A}^{k-1-i}  \mathbf{C} {\mathbf{B}}^{i} =    \sum_{i=0}^{k-1}  \mathbf{A}^{k-1-i} {\mathbf{B}}^{i}$. 
Since $\sigma(\mathbf{A}) \cap  \sigma(\mathbf{B})  =\emptyset $ then $\sigma(\mathbf{A}^{d}) \cap  \sigma(\mathbf{B}^{d})  =\emptyset $. Substituting $\mathbf{A}$ and $\mathbf{B}$ with $\mathbf{A}^{d}$and $\mathbf{B}^{d}$, respectively,
then by similar arguments to Proposition 2.3. in \cite{Poly_Sol_Sylvester} $\mathbf{X}$ can be represented as a polynomial of $\mathbf{A}^{d}$ and $\mathbf{B}^{d}$ using the Cayley–Hamilton Theorem with $\mathbf{X} =  q(\mathbf{A}^{d})^{-1} \eta (k,\mathbf{A}^{d} , \mathbf{B}^{d})$, where $ q(\mathbf{A}^{d})$ is the characteristic polynomial of $\mathbf{A}^{d}$ .  $\square$





\end{document}